\documentclass[runningheads]{llncs}

\newcommand{\fulltask}{Foresight Expression Video Object Segmentation\xspace}
\newcommand{\myparagraph}[1]{{\vspace{.1em} \noindent \bf #1}}

\newcommand{\task}{FeVOS\xspace}
\newcommand{\dataset}{FeVOS\xspace}

\newcommand{\method}{FeVOS-R1\xspace}

\newcommand{\numvideo}{968\xspace}
\newcommand{\numobj}{1,419\xspace}
\newcommand{\nummask}{37,412\xspace}
\newcommand{\numexpre}{14,525\xspace}
\newcommand{\numframe}{30,125\xspace}
\newcommand{\numcot}{2,904\xspace}
\newcommand{\numvideotrain}{779\xspace}
\newcommand{\numvideoval}{189\xspace}
\newcommand{\numexpretrain}{11,708\xspace}
\newcommand{\numexpreval}{2,817\xspace}
\newcommand{\numframetrain}{24,689\xspace}
\newcommand{\numframeval}{5,436\xspace}
\newcommand{\nummasktrain}{30,694\xspace}
\newcommand{\nummaskval}{6,718\xspace}
\newcommand{\numobjtrain}{1,137\xspace}
\newcommand{\numcottrain}{2,337\xspace}
\newcommand{\numobjval}{282\xspace}
\newcommand{\numcotval}{567\xspace}

\newcommand{\cmark}{\ding{51}\xspace}%
\newcommand{\xmarkg}{\textcolor{lightgray}{\ding{55}}\xspace}%

\usepackage{multirow}
\usepackage{pifont}

\usepackage{eccv}

\usepackage{eccvabbrv}
\usepackage[misc]{ifsym}
\usepackage{graphicx}
\usepackage{booktabs}

\usepackage[accsupp]{axessibility}  

\usepackage[pagebackref,breaklinks,colorlinks,citecolor=eccvblue]{hyperref}

\usepackage{orcidlink}

\begin{document}

\title{FeVOS: Foresight Expression Video \\Object Segmentation} 

\titlerunning{FeVOS}

\author{Kehan Lan\orcidlink{0009-0008-4558-7147} \and
Kaining Ying\orcidlink{0000-0003-2596-1847} \and
Henghui Ding\orcidlink{0000-0003-4868-6526}~$^{\textrm{\Letter}}$}

\authorrunning{K.~Lan et al.}

\institute{Institute of Big Data, College of Computer Science and Artificial Intelligence, \\ Fudan University, China\\
\email{\{khlan24, knying24\}@m.fudan.edu.cn, hhding@fudan.edu.cn}\\
\email{\url{https://henghuiding.com/FeVOS/}}}
\begingroup
\renewcommand{\thefootnote}{\textrm{\Letter}}
\footnotetext{Corresponding author}
\endgroup

\maketitle

\begin{abstract}
    Existing Referring Video Object Segmentation tasks focus on referring expressions describing events, actions or appearances of relevant objects within the observed frames, 
    lacking evaluation in scenarios that require pre-decisive spatio-temporal reasoning, thereby limiting their applicability.
    To address this, we propose \textbf{\fulltask}, a task that queries future events in upcoming video segments and requires masks of the objects in the observed frames as visual answers. For example, in ego-centric scenes, the question \textit{``What tool will be used?''} demands reasoning over spatio-temporal cues to predict the masks of the next tool to be used, which helps with the understanding of future actions and decisions.
    To support this task, we introduce \textbf{\dataset}, a dataset with \numvideo video clips, \numexpre foresight expressions, and \numcot chain-of-thought annotations to provide explicit and interpretable reasoning steps. We further develop \textbf{\method}, an MLLM-based model trained on our dataset via a two-stage pipeline of supervised fine-tuning and reinforcement learning. \method not only achieves state-of-the-art performance on \dataset, but also demonstrates strong generalization to existing RVOS benchmarks. 
    We hope this work can inspire more research on predictive reasoning in video perception.

\end{abstract}
        
\section{Introduction}
\label{sec:intro}

Referring Video Object Segmentation (RVOS)~\cite{refer_davis,refer_youtube_vos,MeViS,visa,videolisa,omniavs,mevisv2} is a challenging task that requires vision-language understanding with pixel-level grounding. 
Given a video clip and a referring expression, the model needs to segment the object corresponding to the referring expression throughout the entire video. 
It has shown great potential in fields like video editing~\cite{tilekbay2024expressedit}, autonomous driving~\cite{lin2024echotrack}, and language-guided robotic planning~\cite{zhou2025roborefer, huang2025roboground, stone2023open}. 
However, existing RVOS tasks and datasets focus solely on grounding expressions within observed frames, lacking the capability to anticipate future events, which is a critical requirement for proactive decision-making in real-world applications.

\begin{figure*}[!t]
    \centering
    \captionsetup{type=figure}
    \includegraphics[width=0.999\textwidth]{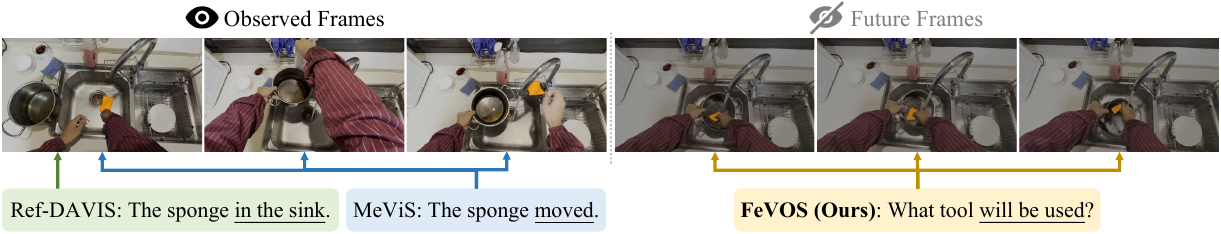}
    \captionof{figure}{\small 
        Comparison of related datasets with \textbf{F}oresight \textbf{e}xpression \textbf{V}ideo \textbf{O}bject \textbf{S}egmentation (\textbf{FeVOS}). Unlike existing datasets (Ref-DAVIS~\cite{refer_davis}, MeViS~\cite{MeViS}) that ground expressions describing \textit{observable} events (\eg, \underline{in the sink}, \underline{moved}), our \task requires predicting which object \textit{will be involved} in future events based on observed visual cues. In this case, given the foresight expression ``\textit{What tool \underline{will be used}?}'', the model must analyze temporal context (dirty pot) and spatial cues (hand states) 
        to anticipate the correct target. Zoom in for a better view.
    }
    \label{fig:teaser}
    \vspace{-5mm}
\end{figure*}

This limitation is reflected in design paradigms of existing RVOS datasets. Earlier RVOS datasets, \eg, Ref-DAVIS~\cite{refer_davis} and Ref-YouTube-VOS~\cite{refer_youtube_vos}, provide videos in various scenarios with diverse object categories, but mainly focus on static attributes of objects (\eg, appearances, categories, positions) that can be inferred from a single frame.
Later, MeViS~\cite{MeViS,mevisv2} was proposed to involve spatio-temporal understanding by introducing motion expressions that describe motion-related attributes of objects across different frames.
However, the referring scope of expressions in these datasets is limited to the \textit{observed} frames provided. As shown in \Cref{fig:teaser}, existing benchmarks like Ref-DAVIS ground expressions such as \textit{``the sponge in the sink''} based on observable attributes, while MeViS handles motion-related expressions like \textit{``the sponge moved''}, both describing events already present in the provided frames.
In contrast, robotic applications often require models to anticipate which object(s) might be involved in the next scene or action before it occurs. This motivates our predictive reformulation of the task, where only visual cues prior to the referred action are provided. In the kitchen scene shown in \Cref{fig:teaser}, the expression \textit{``What tool will be used?''} requires the model to analyze both temporal context and spatial cues in the observed frames to predict which object will be involved in the future cleaning action. In this case, visual evidence of the dirty pot with dish soap applied signals a temporal transition toward cleaning action, allowing the model to infer that a sponge will be used next. Beyond tool category prediction, the model must also perform spatial disambiguation to locate the sponge on the right side, which is more convenient for manipulation since the left hand remains occupied. While existing video understanding~\cite{wang2025fosteringvideoreasoningnextevent, li2024mvbench, liu2024mmbench, yu2024merlin} tasks have explored temporal reasoning and future prediction, they typically follow a simple QA paradigm, lacking the fine-grained pixel-level grounding necessary for actionable decision-making.

To address these limitations, we propose Foresight Expression Video Object Segmentation (FeVOS), a novel task that queries future events in upcoming video segments and requires segmentation masks of the relevant objects in the observed frames. 
To support this task, we introduce \dataset, a carefully curated dataset with \numvideo video clips spanning diverse scenes, \numexpre foresight expressions, and corresponding segmentation masks.
Unlike traditional RVOS or existing video understanding tasks, our task introduces additional challenges: (I) The information provided in the question is limited to the future, making it difficult to directly align text expressions with the currently observed visual content. (II) It requires analyzing both temporal context and spatial cues to predict which object(s) will be involved in future actions, demanding a combination of visual reasoning and world knowledge. (III) The model must accurately provide pixel-level segmentation masks for the relevant object(s) in the observed frames, enriching future prediction with fine-grained spatio-temporal understanding.

While existing MLLM-based methods~\cite{sa2va,visa,videolisa, lin2025glus, gong2025devil} achieve strong performance on traditional benchmarks that involve observable events, they struggle with our predictive reasoning task that requires anticipating future events from observed visual cues. To address this challenge, we draw inspiration from recent advances in reasoning-oriented reinforcement learning~\cite{deepseekai2025deepseekr1incentivizingreasoningcapability} and develop \method, a model trained via a two-stage pipeline. In Stage 1, we perform chain-of-thought supervised fine-tuning as a cold start, leveraging \numcot synthetic CoT annotations we generated for \dataset that provide step-by-step reasoning paths (as shown in \Cref{fig:data_samples}). In Stage 2, we employ Group Relative Policy Optimization (GRPO)~\cite{shao2024deepseekmath} with proper reward, enabling the model to directly optimize for task performance while maintaining interpretable reasoning. Experiments demonstrate that \method achieves state-of-the-art performance on \dataset while generalizing well to existing RVOS benchmarks~\cite{visa,MeViS} without additional fine-tuning.
In a nutshell, our main contributions are as follows:

\begin{figure*}[!t]
  \centering
  \includegraphics[width=0.995\textwidth]{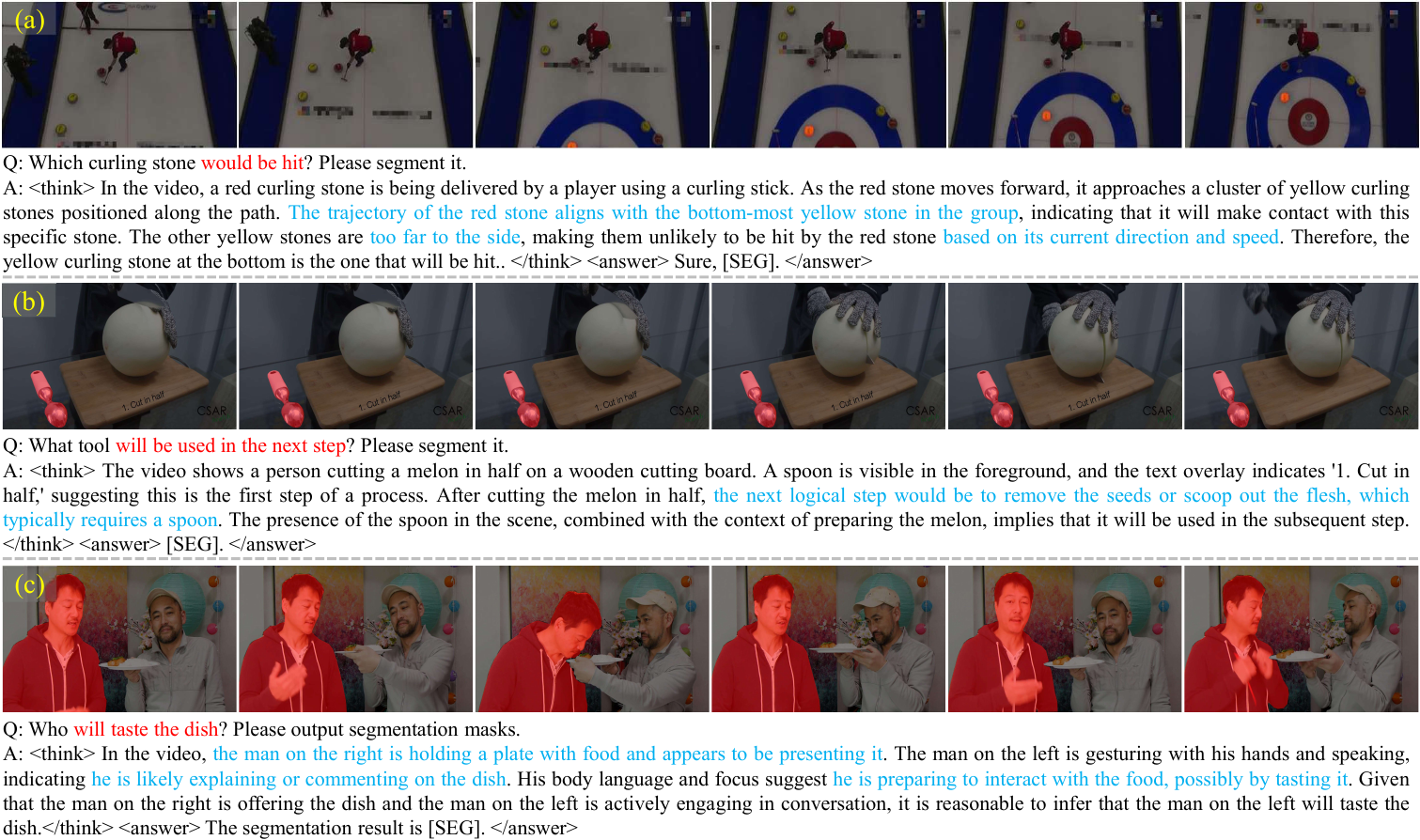}
  \caption{Samples from \dataset with chain-of-thought annotations. These examples present representative reasoning challenges included in our task: (a) Physically-Aligned Prediction, (b) Procedure-Grounded Prediction, (c) Intention-Guided Prediction. \textcolor{red}{Red text} highlights foresight expressions describing future, while \textcolor{cyan}{blue text} indicates key reasoning steps that lead to identifying the target objects. Zoom in for a better view.}
  \label{fig:data_samples}
  \vspace{-4mm}
\end{figure*}

\begin{itemize}
  \item We propose \textbf{\fulltask}, a task requiring prediction of which object(s) will be involved in future events from observed visual cues, advancing RVOS toward predictive reasoning.
  \item We introduce \textbf{\dataset dataset}, containing \numvideo videos, \numexpre foresight expressions, and corresponding pixel-level masks, along with \numcot synthetic chain-of-thought annotations that provide explicit reasoning supervision.
  \item We develop \textbf{\method}, a two-stage training framework combining supervised fine-tuning with CoT data and reinforcement learning with end-to-end rewards, achieving state-of-the-art performance on predictive segmentation while generalizing well to traditional RVOS benchmarks.
\end{itemize}

\section{Related Work}
\label{sec:relatedwork}

\subsection{Referring Video Object Segmentation}
Referring Video Object Segmentation (RVOS)~\cite{refer_davis,refer_youtube_vos,MeViS,visa,videolisa} aims to segment objects corresponding to given language expressions in videos. 
Early RVOS datasets~\cite{refer_youtube_vos,refer_davis,a2d} predominantly focus on static attributes that can be inferred from single frames. 
MeViS~\cite{MeViS} advances the field by introducing motion expressions that require spatio-temporal understanding across frames.
Recent works like ReVOS~\cite{visa} and ReasonVOS~\cite{videolisa} further incorporate reasoning capabilities and world knowledge. 
However, all of these datasets perform reasoning on \textit{observed} frames and focus on grounding expressions describing events already present in the video. In contrast, our \dataset requires models to predict which object(s) \textit{will be involved} in future events based solely on observed visual cues, introducing a fundamentally different predictive reasoning challenge.

\subsection{MLLMs for Segmentation}
As MLLMs~\cite{internvl,chen2024internvl,qwen2vl,qwenvl,qwen2.5vl, llava, chatunivi,mmtbench,convbench} advance in vision-language understanding and reasoning, several works~\cite{lisa,visa,trackgpt,lin2025glus,gong2025devil} uncover their significant potential for visual grounding tasks like referring segmentation.
LISA~\cite{lisa} pioneers the introduction of a special token \texttt{[SEG]} into MLLMs and appends a segmentation head, thereby unleashing their segmentation capabilities for referring image segmentation tasks. VideoLISA~\cite{videolisa} and TrackGPT~\cite{trackgpt} further extend this MLLM-based approach to the video domain.
VISA~\cite{visa} proposes a two-stage pipeline that first localizes the frames where the target object appears and then performs segmentation. Subsequent works~\cite{gong2025devil,lin2025glus} have significantly boosted the performance by refining the frame-selection stage.
Sa2VA~\cite{sa2va} proposes a unified framework that seamlessly integrates an MLLM with SAM2~\cite{sam2}, delivering superior performance on both video understanding and segmentation tasks.

\subsection{Reinforcement Learning for Vision-Language}
Deepseek-R1~\cite{deepseekai2025deepseekr1incentivizingreasoningcapability} demonstrates that Reinforcement Learning~(RL)~\cite{sutton1998reinforcement} effectively enhances LLM reasoning via GRPO~\cite{shao2024deepseekmath}.
VLM-R1~\cite{shen2025vlm} extends the training paradigm to more general vision-language tasks~(referring expression comprehension and open-vocabulary object detection).
SegAgent~\cite{zhu2025segagent} trains MLLMs to mimic the annotation trajectories of human annotators.
Seg-Zero~\cite{liu2025segzero} implements RL training from scratch by asking MLLMs to directly generate the coordinates of bounding boxes to prompt SAM2.
Concurrently, Veason-R1~\cite{gong2025reinforcing} explores RL for video segmentation.
While concurrent work focuses on segmenting objects with expressions in traditional RVOS settings, we focus on predictive reasoning that requires anticipating future events from observed visual cues.

\begin{figure*}[!t]
  \centering
  \includegraphics[width=\textwidth]{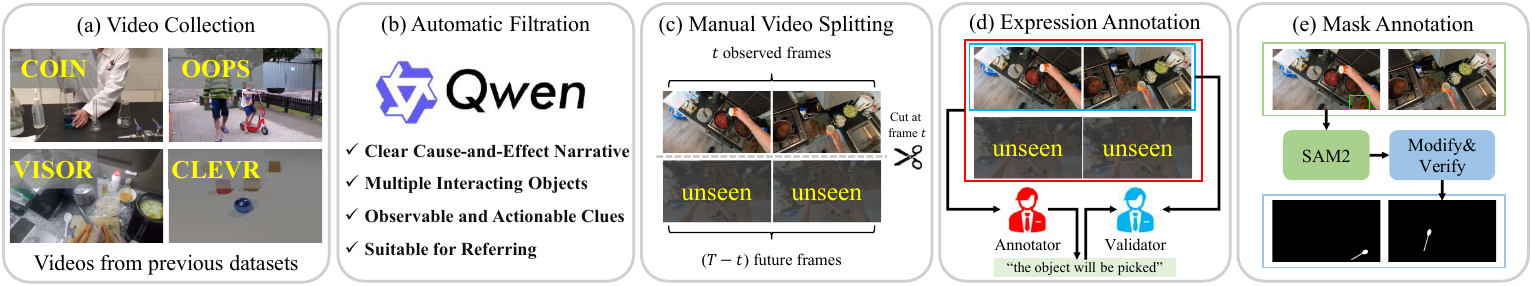}
  \caption{Data Construction Pipeline. (a) Video Collection. Diverse videos were gathered from multiple sources~\cite{coin, star, visor, johnson2017clevr, epstein2020oops}. (b) Automatic Filtration. We used Qwen2.5-VL~\cite{qwen2.5vl} to filter videos based on carefully designed rules. (c) Manual Video Splitting. We manually split the videos that meet our criteria to observed frames and future frames. (d) Expression Annotation. We designed foresight expressions through annotation and validation to minimize the ambiguity while keeping the task challenging. (e). Mask Annotation. We used an interactive tool to annotate masks in the observed frames.}
  \label{fig:anno_ppl}
  \vspace{-2mm}
\end{figure*}

\section{Benchmark: \dataset}

\subsection{Task Formulation}
Given a video clip $V \in \mathbb{R}^{T \times 3 \times H \times W}$, where $T$, $H$, and $W$ denote the length, height, and width of frames, and a predictive expression $Q$ describing \textit{future events}, the model outputs pixel-level segmentation masks $M \in \mathbb{R}^{T \times H \times W}$ for objects in the observed frames. Unlike traditional RVOS, our task requires reasoning about future actions or events based solely on cues from observed frames.

\subsection{Construction of \dataset}
\label{sec:data_construction}
We demonstrate the five-stage \dataset construction pipeline in \Cref{fig:anno_ppl} and describe each stage in detail below. 

\myparagraph{(a) Video Collection.} 
We collected candidate videos from various established benchmarks, including COIN~\cite{coin}, STAR~\cite{star}, CLEVR~\cite{johnson2017clevr}, OOPS~\cite{epstein2020oops}, and EPIC-KITCHENS-VISOR~\cite{visor}. These datasets cover diverse scenarios such as egocentric activities, instructional videos, and dynamic outdoor environments. They are well-suited for our predictive reasoning and grounding task as they exhibit rich spatio-temporal dynamics and causal relationships, where early visual cues could naturally foreshadow subsequent events.

\myparagraph{(b) Automatic Filtration.} 
We employed Qwen2.5-VL~\cite{qwen2.5vl} to automatically filter videos for predictive suitability. The model evaluated whether each video could be temporally divided into two clips where: (1) the first clip contains a clear cause-and-effect narrative that enables deterministic inference of events in the second clip; (2) multiple interacting objects or agents are present; (3) the scene contains observable and actionable clues; and (4) the video is suitable for designing referring expressions. Around 30.5\% videos are retained in this stage.

\myparagraph{(c) Manual Video Splitting.}
We designed an online tool which can be used to load videos and decide to label time points at 0.2-second intervals or discard a video. Expert annotators manually reviewed the filtered videos and determined the optimal temporal split points for each video. The split point was chosen to ensure that: (1) the observation clip contains visible target objects with implicit visual cues indicating their future involvement; (2) the future clip contains clear, predictable events or actions; and (3) the split maximizes the predictive challenge while maintaining reasonable inference. Videos that could not be meaningfully split according to these criteria were excluded (around 44.0\% retained).

\myparagraph{(d) Expression Annotation.} 
To ensure high-quality predictive expressions, we adopted a two-stage annotation process. In the first stage, annotators with access to both observed and unseen clips selected target objects in the observed frames and designed expressions describing future events, \eg, \textit{``the object that will be picked''}. In the second stage, independent validators, provided with the expression and observed frames without access to unseen clips, attempted to identify the target objects through predictive reasoning. Crucially, expressions must describe future events rather than directly describing observable attributes in the current frames. Around 85.9\% of expressions are retained to ensure that they both required genuine predictive reasoning and allowed validators to correctly identify targets. This two-stage process ensures that (1) our dataset truly demands anticipatory reasoning rather than simple object recognition and (2) enough information is provided to predict the answers. Notably, although the future has its ambiguity and there might be multiple plausible outcomes in real-world scenarios, we only retain video-expression pairs with target object(s) that can be reasonably inferred with observable spatio-temporal cues. This can be further ensured in this expression annotation stage by adding restrictions (\eg ``first'' when the referred events might happen multiple times).

\myparagraph{(e) Mask Annotation.} 
We generated pixel-level segmentation masks using an interactive annotation tool built upon SAM2~\cite{sam2}. Annotators carefully modified and verified the masks of the target objects across all frames in the observed part, ensuring spatial and temporal consistency.

\subsection{Automatic CoT Annotation Generation}
\label{sec:data_cot}

After obtaining the video-object-expression triplets through the aforementioned pipeline, we generated chain-of-thought annotations to enable explicit reasoning processes. Specifically, we leveraged Qwen2.5-VL~\cite{qwen2.5vl} to automatically produce CoT annotations using visual prompting. For each triplet, we overlaid the ground-truth segmentation masks onto the video frames as visual prompts, highlighting the target objects. The model was then provided with both the masked video and the foresight expression, and prompted to generate step-by-step reasoning that explains why the highlighted object is the answer to the given predictive question. This reasoning chain typically includes analysis of visual cues, temporal context, and causal relationships observed in the frames. We generated 3 CoT annotations per video to provide diverse reasoning perspectives. As shown in \Cref{fig:data_samples}, the CoT annotations demonstrate how to analyze temporal context and causal relationships to identify target objects that will participate in future events. This automatic annotation process enables our model to learn interpretable reasoning patterns that bridge the gap between visual observations and predictive conclusions and helps with the reinforcement learning stage.

\begin{figure}[!t]

\begin{center}

\begin{minipage}[!t]{0.32\textwidth}
    \centering
    \includegraphics[width=\textwidth]{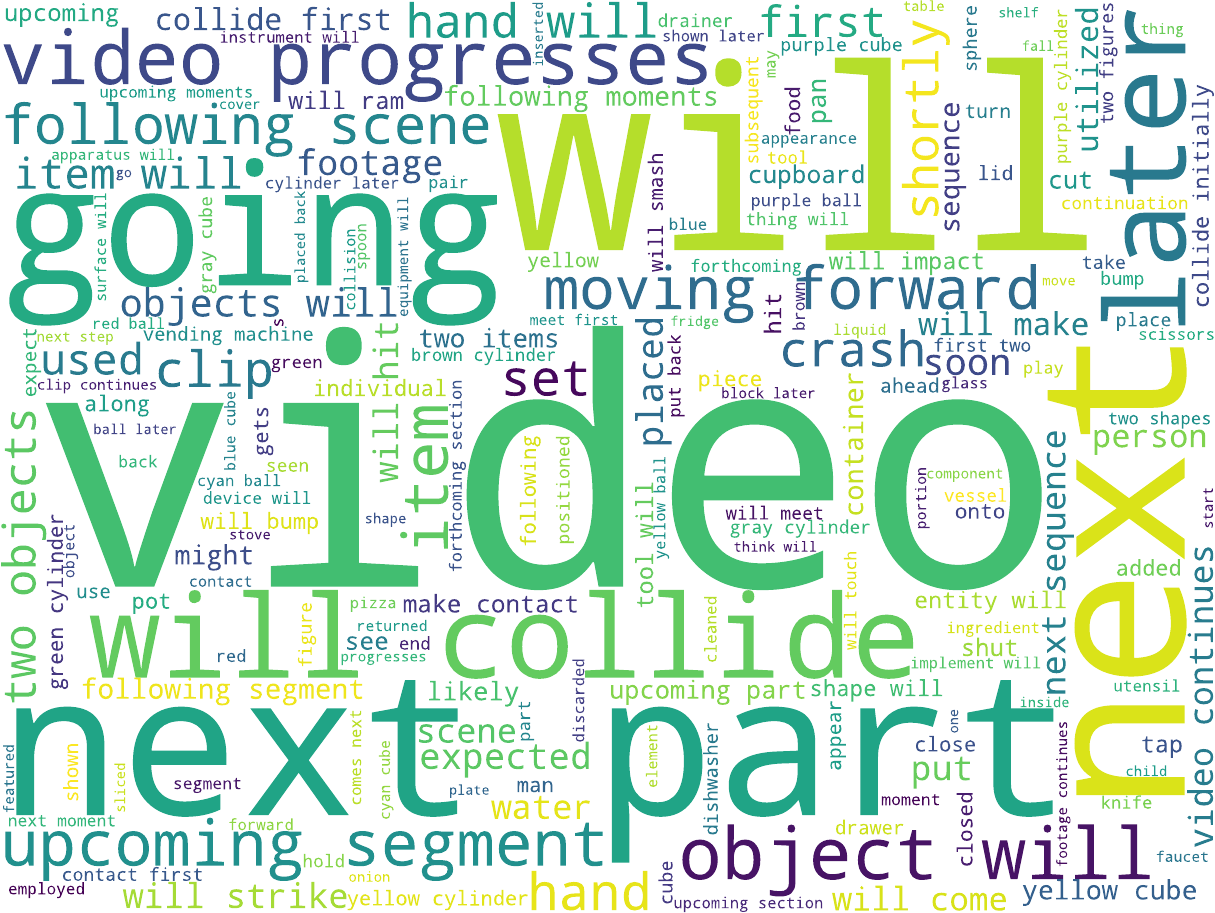}
    \captionof{figure}{Expressions cloud.}
    \label{fig:wc_exp}
\end{minipage}
\hfill
\begin{minipage}[!t]{0.32\textwidth}
    \centering
    \includegraphics[width=\textwidth]{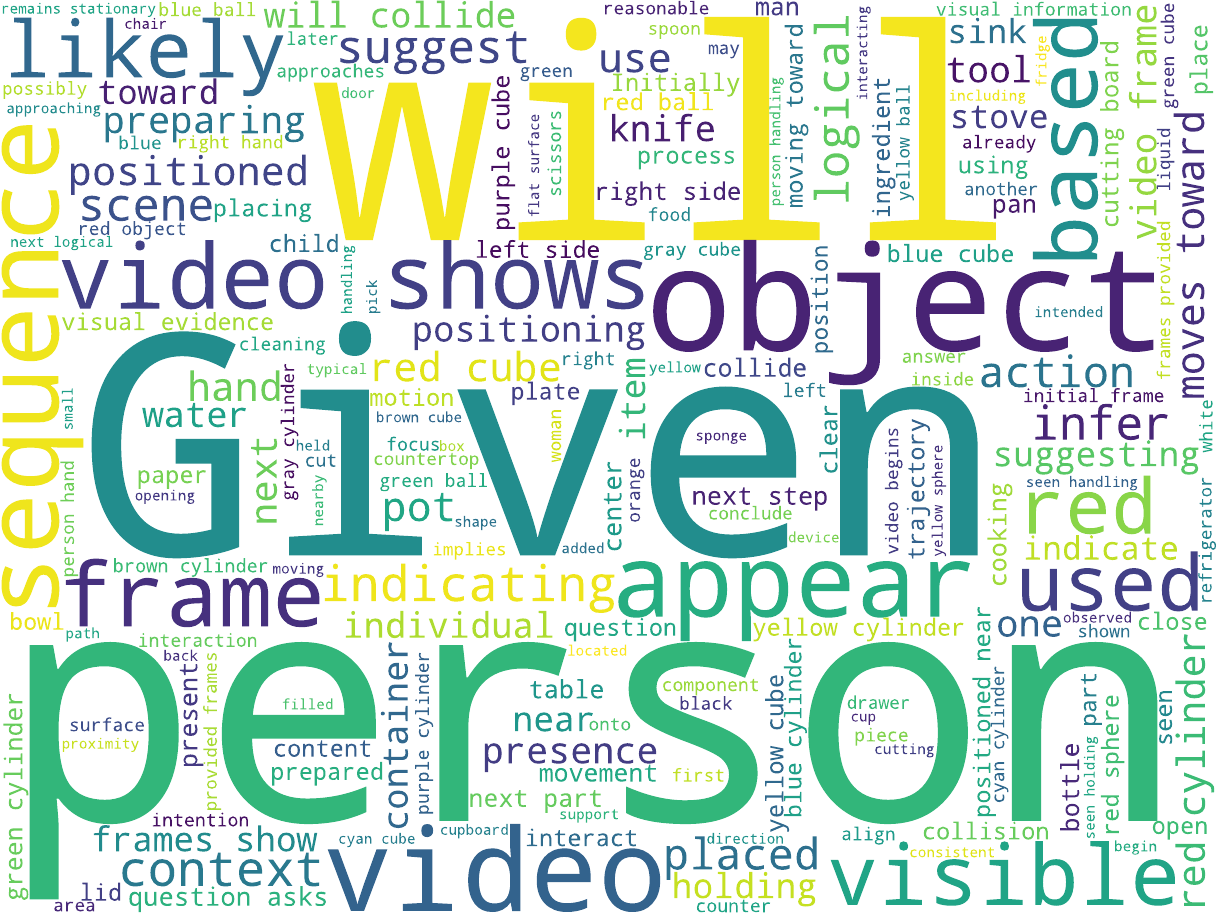}
    \captionof{figure}{Reasoning cloud.}
    \label{fig:wc_reasoning}
\end{minipage}
\hfill
\begin{minipage}[!t]{0.32\textwidth}
    \centering
    \small
    \setlength{\tabcolsep}{0.8pt}
    \renewcommand{\arraystretch}{1.0}
    \begin{tabular}{l|c|c|c} 
        \toprule
        Stat. & Train & Val & All \\
        \hline
        Videos  & \numvideotrain & \numvideoval & \numvideo \\
        Expr.   & \numexpretrain & \numexpreval & \numexpre \\
        Frames  & \numframetrain & \numframeval & \numframe \\
        Obj.    & \numobjtrain & \numobjval & \numobj \\
        Masks   & \nummasktrain & \nummaskval & \nummask \\
        CoT     & \numcottrain & \numcotval & \numcot \\
        \bottomrule
    \end{tabular}
    \captionof{table}{\dataset statistics.}
    \label{tab:stats}
\end{minipage}
\vspace{-4mm}
\end{center}
\end{figure}

\subsection{Dataset Statistics \& Analysis}
Through the aforementioned annotation pipeline, we curated \dataset, which contains \numvideo video clips with \numexpre foresight expressions and corresponding pixel-level segmentation masks. Each expression is paired with precise annotations identifying target objects across all frames in the observation segment. Additionally, we generated \numcot synthetic chain-of-thought annotations to provide explicit reasoning supervision. 
We split \dataset into training and validation subsets. The detailed dataset statistics can be found in ~\Cref{tab:stats}.

\myparagraph{Word Cloud.} We provide word clouds illustrating the distribution of grounding expressions and reasoning processes in~\Cref{fig:wc_exp,fig:wc_reasoning}. Notably, our foresight expressions frequently contain phrases like \textit{``the next part of the video''} and words like \textit{``will''}, which are intentionally used to encourage models to anticipate forthcoming events and actions before segmenting. Additionally, in reasoning processes, terms like \textit{``Given''} appear frequently, guiding the model to align its inferences with the visual evidence provided by observed frames.

\myparagraph{Predictive Reasoning Challenges.} Compared with former RVOS datasets~\cite{MeViS,refer_youtube_vos,refer_davis}, our task needs the model to perform predictive saptio-temporal reasoning. We further analyze the three representative examples in~\Cref{fig:data_samples} to illustrate the diverse reasoning challenges posed by our task including: (a) Physically-Aligned Prediction. The model must use the cross-frame information to predict the trajectory of the red curling stone to identify the right yellow curling stone that will be hit. (b) Procedure-Grounded Prediction. The model must understand the current cutting operation and reason about the whole logical process in the procedure to decide that the spoon would be involved in the next step in removing the flesh. (c) Intention-Guided Prediction. The model should distinguish between the presenter and the potential taster, aligning the tasting action with the correct individual by inferring their intentions through their interactions. These challenges require a combination of spatio-temporally grounded reasoning and world knowledge to make the right mask prediction.

\subsection{Evaluation Metrics}
Following standard practice in video object segmentation \cite{MeViS,visa,MOSEv2,move,saas}, we employ two complementary metrics: $\mathcal{J}$ (region similarity via IoU) and $\mathcal{F}$ (boundary accuracy), and use their mean $\mathcal{J}\&\mathcal{F}$ as the overall performance metric.

\section{Baseline: \method}

\subsection{Preliminary}

\myparagraph{Sa2VA.}
Sa2VA~\cite{sa2va} is a unified framework that integrates MLLM~\cite{internvl} with SAM2~\cite{sam2} for referring video object segmentation. The architecture consists of three key components: (1) a vision encoder that extracts visual features from video frames, (2) a large language model that processes both visual embeddings and text prompts to generate responses with special segmentation tokens, and (3) SAM2's mask decoder that produces pixel-wise segmentation masks conditioned on the hidden states of segmentation tokens. By leveraging the reasoning capabilities of LLMs and the strong segmentation performance of SAM2, Sa2VA achieves state-of-the-art results on traditional RVOS benchmarks. However, its supervised fine-tuning paradigm struggles with predictive reasoning tasks that require anticipating future events from observed visual cues, as it lacks explicit reasoning processes that lead to better optimization for segmentation quality. To address this limitation, we propose a two-stage training paradigm that equips the model with interpretable chain-of-thought reasoning processes and aligns it with task-specific objectives through reinforcement learning.

\myparagraph{Group Relative Policy Optimization.}
GRPO~\cite{shao2024deepseekmath} is a value-free reinforcement learning algorithm for efficient policy optimization. For each query $q$, GRPO samples a group of outputs $G=\{o_i\}_{i=1}^{|G|}$ and computes their rewards $\{r_i\}$. The advantage function is normalized relative to group statistics: $A_i = (r_i - \bar{r}_G) / \sigma_G$, where $\bar{r}_G$ and $\sigma_G$ denote the mean and standard deviation of rewards. The policy $\pi_\theta$ is updated by maximizing the following objective:
\begin{equation}
    \mathcal{J}(\theta)\! = \! 
    \mathbb{E}_{G} \!\!\left[\!
        \frac{1}{|G|} \!
        \sum_{i \in G} \!\left(
            \min\left(
                s_i A_i,
                \operatorname{clip}\!\big(s_i, 1-\epsilon, 1+\epsilon\big) A_i
            \right)\!-\!\beta\mathbb{D}_\mathrm{KL}(\pi_\theta || \pi_{ref}) 
        \right)\!
    \right],
    \label{eq:grpo}
\end{equation}
where $s_i=\frac{\pi_\theta(o_i|q)}{\pi_{\text{old}}(o_i|q)}$ denotes the probability ratio between the new and old policies, and $\epsilon$ is the clipping parameter controlling the update range. $\mathbb{D}_\mathrm{KL}(\pi_\theta || \pi_{ref})$ is a KL divergence regularization term with a coefficient $\beta$ that penalizes the model from deviating excessively from a reference model $\pi_{ref}$.

\begin{figure*}[!t]
  \centering
  \includegraphics[width=\textwidth]{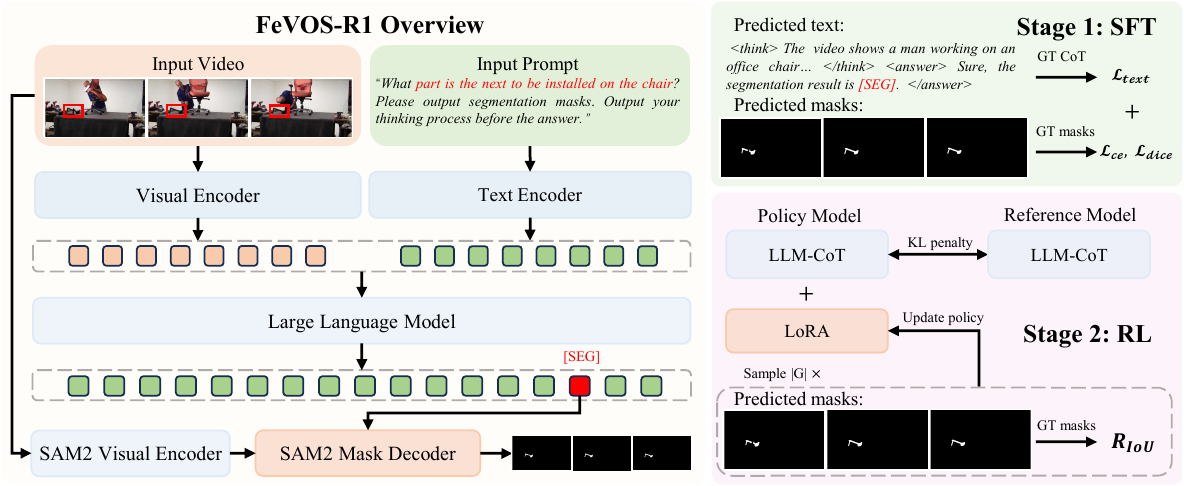}
  \caption{\textbf{Overview of \method.} Our two-stage training framework: Stage 1 performs SFT with chain-of-thought annotations, and Stage 2 employs GRPO with IoU-based rewards to optimize segmentation quality through reasoning optimization.}
\label{fig:method}
\end{figure*}

\subsection{Two-Stage Training Pipeline}
  
We implement our method \method based on Sa2VA via a two-stage training paradigm. As shown in \Cref{fig:method}, given an input video $V$, we first sample a sequence of frames $\{I_t\}_{t=1}^{T}$ and encode them using the vision encoder to obtain visual embeddings. These embeddings, along with a text prompt $P$, are processed by the LLM to generate a response containing a special token \texttt{[SEG]}. The hidden states of \texttt{[SEG]} are then projected and fed into SAM2's mask decoder to predict a segmentation mask sequence $\{\hat{M}_t\}_{t=1}^{T}$ for the input frames.

\myparagraph{Stage~1:~Supervised Fine-Tuning with Chain-of-Thought.}~To equip the model with basic reasoning capabilities, we perform supervised fine-tuning using our synthetic CoT dataset introduced in Section~\ref{sec:data_cot}. During training, the model learns to generate step-by-step reasoning processes that analyze temporal context and causal relationships before producing the \texttt{[SEG]} token. We employ three loss functions: pixel-wise cross-entropy loss $\mathcal{L}_{ce}$ and Dice loss $\mathcal{L}_{dice}$ for evaluating mask quality, and a text generation loss $\mathcal{L}_{text}$ for supervising the reasoning processes. The total loss is formulated as:

\begin{equation}
  \mathcal{L}_{total} = 
  \alpha_{ce}\mathcal{L}_{ce} + \alpha_{dice}\mathcal{L}_{dice} + \alpha_{text}\mathcal{L}_{text},
  \label{eq:loss}
\end{equation}

where $\alpha_{ce}, \, \alpha_{dice}, \, \alpha_{text}$ represent the weights of the three losses, respectively. This stage enables the model to generate interpretable reasoning in the correct format while producing segmentation masks, laying the foundation for subsequent training with reinforcement learning.

\myparagraph{Stage 2: Reinforcement Learning with End-to-End Rewards.}
While supervised fine-tuning provides preliminary knowledge of reasoning, the reasoning process remains suboptimal in quality and weakly aligned with segmentation objectives. We employ GRPO to further refine the reasoning process that leads to high segmentation quality with task-specific objectives. Unlike existing visual grounding methods~\cite{liu2025segzero, gong2025reinforcing} that require models to output intermediate representations (\eg, bounding box coordinates in JSON format), we leverage the \texttt{[SEG]} token to support end-to-end optimization that directly maximizes segmentation accuracy. Specifically, we define an accuracy reward $R_{IoU}$ from the IoU between predicted and ground-truth masks:
\begin{equation}
R_{IoU} = \frac{1}{T}\sum_{t=1}^{T}\text{IoU}(\hat{M}_t, M_t),
  \label{eq:acc_reward}
\end{equation}
where $\hat{M}_t$ and $M_t$ denote the predicted and ground-truth masks at frame $t$, respectively. While previous works \cite{deepseekai2025deepseekr1incentivizingreasoningcapability, liu2025segzero, gong2025reinforcing} typically incorporate a format reward to enforce structured output with \texttt{<think></think>} and \texttt{<answer></answer>} tags and correct JSON format, we find that using the accuracy reward alone achieves better performance. Our ablation studies in Section~\ref{sec:ablation} reveal that format constraints provide no additional benefit and can even impede optimization. This is largely because our reasoning format is lightweight and already well-learned in the SFT stage. Consequently, our GRPO stage focuses exclusively on segmentation-aware reasoning refinement, guided solely by $R_{IoU}$, enabling tighter alignment between reasoning quality and segmentation performance.
\section{Experiments}
\label{sec:exp}

\subsection{Implementation Details}
\myparagraph{Supervised Fine-Tuning.}
This stage serves as the CoT cold start before RL.
We adopt pretrained Sa2VA-4B~\cite{sa2va} as our base model, which integrates InternVL2.5~\cite{internvl} as the MLLM and SAM2-L~\cite{sam2} as the segmentation module. 
During training, we only update the LLM and the SAM2 mask decoder while keeping other parts frozen, and apply LoRA~\cite{lora} (rank = 128) for efficient parameter tuning.
The optimization is performed using a learning rate of $2 \times 10^{-5}$ with the cosine annealing schedule. We set the batch size to 4 with gradient accumulation over 4 steps and train on \dataset dataset with CoT annotations for 4 epochs.

\myparagraph{Reinforcement Learning.} We freeze SAM2 mask decoder and exclusively fine-tune the LLM component using the same LoRA configuration as in the SFT stage. During training, the model generates $|G| = 4$ responses per input for GRPO. We set the learning rate to $1 \times 10^{-5}$ with a batch size of 4 and gradient accumulation over 2 steps, training on the same expressions as SFT stage for 2 epochs. All experiments are conducted on 4 NVIDIA RTX 4090 GPUs.

\myparagraph{Inference.} We get a CoT response followed by a segmentation answer from the model and prompt SAM2 with the output \texttt{[SEG]} token to get final masks.

\subsection{Quantitative Results}

\begin{table}[t]
    \centering
    \small
    \renewcommand\arraystretch{1.0}
    \setlength{\tabcolsep}{11.9pt}
    \caption{
    Main results on the \dataset dataset. 
    * indicates the model is fine-tuned on \dataset. 
    \textbf{Bold} indicates the best performance.
}
    \label{table:fevos_results}
        \begin{tabular}{l|l|ccc}
        \toprule
        Method & Backbone & $\mathcal{J}$ & $\mathcal{F}$ & $\mathcal{J}\&\mathcal{F}$ \\
        \hline

        \multicolumn{5}{l}{\emph{Zero-shot baselines}} \\
        \hline
        ReferFormer~\cite{referformer} & ResNet-50 & 16.4 & 20.0 & 18.2 \\
        LMPM~\cite{MeViS} & Swin-T & 17.1 & 20.7 & 18.9 \\
        VISA~\cite{visa} & Chat-UniVi-7B & 22.9 & 28.2 & 25.6 \\ 
        VideoLISA~\cite{videolisa} & LLaVA-Phi-3-V-3.8B & 22.9 & 29.4 & 26.1\\
        VideoGLaMM~\cite{munasinghe2025videoglamm} & Phi3-Mini-3.8B & 21.7 & 26.7 & 24.2 \\
        VRS-HQ~\cite{gong2025devil} & Chat-UniVi-7B & 28.8 & 33.3 & 31.0 \\
        GLUS~\cite{lin2025glus} & Chat-UniVi-7B & 27.4 & 31.7 & 29.6 \\
        Sa2VA~\cite{sa2va} & InternVL2.5-4B & 23.7 & 27.2 & 25.4 \\
        \hline
        \multicolumn{5}{l}{\emph{Fine-tuned on \dataset}} \\
        \hline
        GLUS*~\cite{lin2025glus} & Chat-UniVi-7B & 31.0 & 35.9 & 33.5 \\
        Sa2VA*~\cite{sa2va} & InternVL2.5-4B & 33.1 & 38.4 & 35.8 \\
        \textbf{\method (ours)} & InternVL2.5-4B & \textbf{39.5} & \textbf{45.1} & \textbf{42.3} \\
        \bottomrule
    \end{tabular}
     \vspace{-3mm}
\end{table}

\myparagraph{Results on \dataset.}
As shown in \Cref{table:fevos_results}, we comprehensively benchmark a range of recent video segmentation models on the proposed \dataset dataset.
Zero-shot models exhibit substantial difficulty with our predictive reasoning task, with $\mathcal{J}\&\mathcal{F}$ scores below 31.0. Traditional RVOS methods such as ReferFormer (18.2) and LMPM (18.9) struggle significantly, as they are designed for grounding explicit descriptions rather than anticipating future events. Recent video-based MLLMs including VideoLISA (26.1) and VideoGLaMM (24.2) show moderate improvements through stronger video-language understanding. Models with advanced reasoning capabilities, particularly VRS-HQ (31.0) and GLUS (29.6), achieve the best zero-shot performance, yet remain substantially below fine-tuned models.
When directly fine-tuning Sa2VA on \dataset using standard SFT without CoT enhancement, performance improves markedly from 25.4 to 35.8, representing a substantial gain of +10.4. While fine-tuning GLUS yields a +3.9 performance gain, it continues to underperform relative to fine-tuned Sa2VA. This confirms that domain-specific adaptation enables models to better capture the temporal dynamics and causal relationships in predictive scenarios, and Sa2VA better learns these patterns and thus serves as our primary baseline. Our complete training pipeline, incorporating both CoT-guided reasoning and RL-based optimization, further pushes the performance to 42.3, achieving an additional +6.5 gain over the SFT baseline. This demonstrates that explicit reasoning chains and reward-guided optimization are essential for capturing subtle visual cues required for accurate future event prediction. Notably, the performance on \dataset (42.3 $\mathcal{J}\&\mathcal{F}$) is substantially lower than on ReVOS (60.3) and MeViS (49.5), highlighting the increased complexity and challenge posed by our predictive segmentation task, which requires models to anticipate future events from visual cues rather than grounding expressions about observable events.

\myparagraph{Generalization to Related Benchmarks.}
To assess generalization capability, we evaluate on ReVOS and MeViS benchmarks in \Cref{table:combined_results}. 
On ReVOS, the directly fine-tuned baseline Sa2VA* suffers a performance drop from 59.1 to 58.1 compared to its zero-shot counterpart, suggesting overfitting to \dataset, while our \method achieves 60.3, outperforming both baselines with particularly strong gains on the reasoning subset (57.8 vs. 55.2 for baseline). On MeViS, our method reaches 49.5, representing a +3.0 improvement over the baseline (46.5), and surpassing all comparable-sized models including VideoLISA (44.4) and VideoGLaMM (45.2), though slightly below larger 7B models like GLUS (51.3) due to model scale differences. These results demonstrate that our CoT-augmented and RL-enhanced training strategy not only improves in-domain performance but also substantially enhances cross-domain generalization, particularly in reasoning-intensive scenes (\eg Reasoning subset of ReVOS).

\begin{table}[t]
    \centering
    \small
    \renewcommand\arraystretch{1.0}
    \setlength{\tabcolsep}{3.7pt}
    \caption{Quantitative results on ReVOS (Referring, Reasoning and Overall $\mathcal{J}\&\mathcal{F}$) and MeViS datasets. "-" denotes results not reported in the original papers.}
    \label{table:combined_results}
    \begin{tabular}{l|l|ccc|ccc}
        \toprule
        \multirow{2}{*}{Method} & \multirow{2}{*}{Backbone} & \multicolumn{3}{c|}{ReVOS} & \multicolumn{3}{c}{MeViS} \\
        & & Ref. & Reas. & All & $\mathcal{J}$ & $\mathcal{F}$ & $\mathcal{J}\&\mathcal{F}$ \\
        \hline
        ReferFormer~\cite{referformer} & ResNet50 & 16.9 & 12.8 & 14.9 & - & - & - \\
        ReferFormer~\cite{referformer} & Video-Swin-B & 32.7 & 23.4 & 28.1 & 29.8 & 32.2 & 31.0 \\
        LMPM~\cite{MeViS} & Swin-T & 34.1 & 18.8 & 26.4 & 34.2 & 40.2 & 37.2 \\
        LISA~\cite{lisa} & LLaVA-7B & 45.7 & 36.1 & 40.9 & 35.1 & 39.4 & 37.2 \\
        TrackGPT~\cite{trackgpt} & LLaVA-7B & - & - & - & 37.6 & 42.6 & 40.1 \\
        VISA~\cite{visa} & Chat-UniVi-7B & 50.9 & 43.0 & 46.9 & 40.7 & 46.3 & 43.5 \\
        VISA~\cite{visa} & LLaVA-7B & 51.0 & 43.2 & 47.1 & - & - & - \\
        VideoLISA~\cite{videolisa} & LLaVA-Phi-3-V-3.8B & - & - & - & 41.3 & 47.6 & 44.4 \\
        VideoGLaMM~\cite{munasinghe2025videoglamm} & Phi3-Mini-3.8B & - & - & - & 42.1 & 48.2 & 45.2 \\
        VRS-HQ~\cite{gong2025devil} & Chat-UniVi-7B & 62.1 & 56.1 & 59.1 & 47.6 & 53.7 & 50.6 \\
        GLUS~\cite{lin2025glus} & Chat-UniVi-7B & 58.3 & 51.4 & 54.9 & \textbf{48.5} & \textbf{54.2} & \textbf{51.3} \\
        Sa2VA~\cite{sa2va} & InternVL2.5-4B & 62.5 & 55.6 & 59.1 & - & - & 46.4 \\
        Sa2VA*~\cite{sa2va} & InternVL2.5-4B & 61.0 & 55.2 & 58.1 & 43.4 & 49.7 & 46.5 \\
        \hline
        \textbf{\method (ours)} & InternVL2.5-4B & \textbf{62.8} & \textbf{57.8} & \textbf{60.3} & 46.2 & 52.7 & 49.5 \\
        \bottomrule
    \end{tabular}
\end{table}
\subsection{Ablation Studies}
\label{sec:ablation}

\begin{table}[t]
    \centering
    \small
    \begin{minipage}{0.48\textwidth}
        \centering
        \setlength{\tabcolsep}{3.6pt}
        \caption{Ablation on training stages.}
        \label{table:ablation_training}
        \begin{tabular}{c|cc|ccc}
            \toprule
            \multirow{2}{*}{ID} & \multicolumn{2}{c|}{Training Stages} & \multicolumn{3}{c}{FeVOS} \\
            & CoT-SFT & RL & $\mathcal{J}$ & $\mathcal{F}$ & $\mathcal{J\&F}$ \\
            \hline
            I & \cmark & \xmarkg & 34.7 & 39.8 & 37.2 \\ 
            II & \xmarkg & \cmark & 33.5 & 38.6 & 36.0 \\ 
            III & \cmark & \cmark & \textbf{39.5} & \textbf{45.1} & \textbf{42.3} \\ 
            \bottomrule
        \end{tabular}
    \end{minipage}
    \hfill
    \begin{minipage}{0.48\textwidth}
        \centering
        \setlength{\tabcolsep}{4.1pt}
        \caption{Ablation on rewards.}
        \label{table:ablation_rewards}
        \begin{tabular}{c|cc|ccc}
            \toprule
            \multirow{2}{*}{ID} & \multicolumn{2}{c|}{Reward} & \multicolumn{3}{c}{FeVOS} \\
            & IoU & Format & $\mathcal{J}$ & $\mathcal{F}$ & $\mathcal{J\&F}$ \\
            \hline
            I & \xmarkg & \cmark & 35.1 & 40.3 & 37.7 \\ 
            II & \cmark & \cmark & 38.3 & 43.5 & 40.9 \\ 
            III & \cmark & \xmarkg & \textbf{39.5} & \textbf{45.1} & \textbf{42.3} \\ 
            \bottomrule
        \end{tabular}
    \end{minipage}
\end{table}
\myparagraph{Two-stage Training Pipeline.} To endow the model with reasoning capabilities and ensure correct output formatting, we adopt a two-stage training pipeline. In the first stage, the model is fine-tuned on \dataset with CoT annotation using supervised learning to acquire basic reasoning skills and learn the expected output structure. In the second stage, we employ GRPO to further refine the model's reasoning chain to progressively enhance its segmentation performance. 
To evaluate the effectiveness of our pipeline, we compare three training strategies as shown in \Cref{table:ablation_training}: (I) SFT-only training with CoT data, (II) pure RL training from scratch, and (III) our proposed two-stage pipeline. SFT-only (37.2 $\mathcal{J\&F}$) outperforms pure RL (36.0 $\mathcal{J\&F}$), validating that explicit reasoning supervision provides a strong foundation. Pure RL struggles as the model produces trivial responses without meaningful reasoning trajectories. Our two-stage pipeline achieves 42.3 $\mathcal{J\&F}$, representing substantial gains of +5.1 over SFT-only and +6.3 over RL-only, demonstrating that the two stages are highly complementary: SFT establishes reasoning patterns and output formatting, while RL further optimizes the reasoning chain through reward-guided exploration and exploitation.

\myparagraph{Reward Design.}
Previous works~\cite{shen2025vlm, liu2025segzero, gong2025reinforcing} using GRPO for vision-language reasoning tasks typically rely on explicit format rewards to encourage structured reasoning outputs and ensure models adhere to specific JSON formats. This design is considered crucial for guiding models in generating proper reasoning processes before structured answers. However, we observe that since our method does not need JSON formats and the reasoning format can be well-learned in SFT, format rewards may become redundant in our experiment setting. To validate this hypothesis, we design a format reward that enforces structured outputs properly separated with reasoning and answer tags. We compare three different reward configurations: IoU reward only, format reward only, and their combination. As shown in \Cref{table:ablation_rewards}, training with the IoU reward alone achieves the best overall performance (42.3 $\mathcal{J\&F}$), outperforming the joint reward setting by 1.4 points and the format reward alone by 4.6 points. This reveals that when our model is properly initialized with basic knowledge of the output format through SFT with CoT data, explicit format rewards become unnecessary and may even divert valuable model capacity from learning task-relevant objectives.

\subsection{Qualitative Results}
\begin{figure*}[!t]
  \centering
  \includegraphics[width=0.99\textwidth]{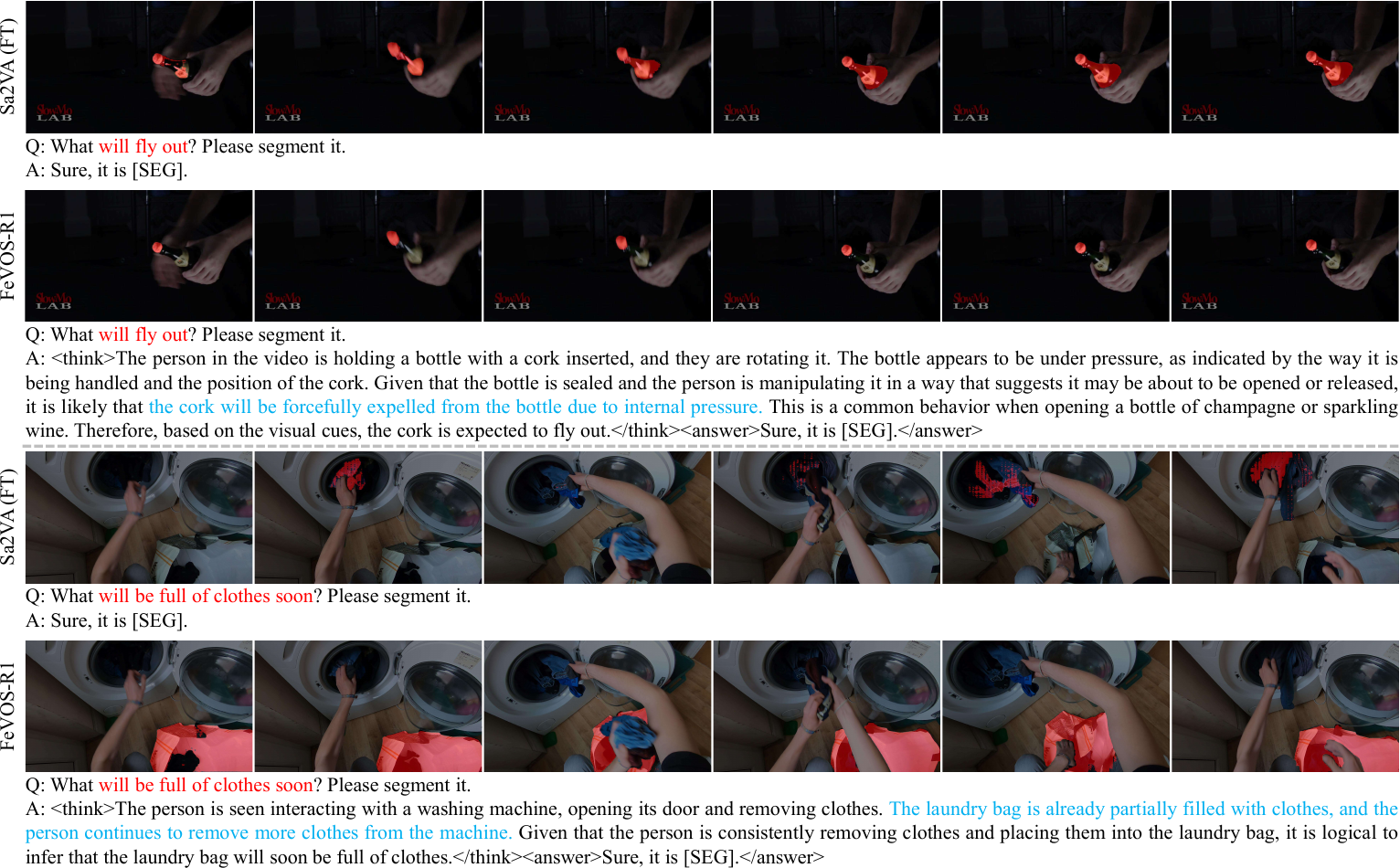}
\vspace{1mm}
  \caption{\textbf{Qualitative comparison.} Our \method generates explicit reasoning with key components (\textcolor{cyan}{blue text}) to analyze visual cues and predict future events, leading to more accurate segmentation results compared to the finetuned baseline Sa2VA.}
  \label{fig:qualitative}
  \vspace{-4mm}
    
\end{figure*}

\Cref{fig:qualitative} shows qualitative comparisons demonstrating the benefit of explicit reasoning. In both cases, the finetuned baseline Sa2VA produces incorrect predictions by failing to analyze temporal context and causal relationships. In contrast, our \method generates detailed reasoning chains that identify key visual cues and predict future events. For example, when asked \textit{``What will fly out?''}, the baseline incorrectly segments the entire bottle, while our method reasons about internal pressure and cork behavior to correctly identify the cork. Similarly, for \textit{``What will be full of clothes soon?''}, our model analyzes the ongoing action of removing clothes to accurately predict the laundry bag. These results demonstrate that explicit reasoning enables more accurate predictions aligned with human intuition and enhances the interpretability.

\section{Conclusion and Discussion}

We introduce \fulltask, a novel task that advances pixel-level video understanding from observation to anticipation through predictive reasoning.
\task requires models to predict future events and segment relevant objects based on implicit visual cues, posing fundamentally new challenges in spatio-temporal reasoning and visual grounding.
To support this task, we construct \dataset, containing \numvideo video clips with \numexpre predictive expressions and corresponding pixel-level segmentation masks, along with \numcot synthetic chain-of-thought annotations to enable explicit reasoning. 
We further develop \method, a reasoning-enhanced model trained through a two-stage pipeline combining supervised fine-tuning with CoT data and reinforcement learning via GRPO. Experiments demonstrate that \method achieves strong performance on \task and exhibits robust generalization to existing RVOS benchmarks. 
Despite significant progress, 
the absolute performance on \dataset remains substantially lower than on traditional RVOS benchmarks, underscoring the inherent difficulty of predictive reasoning and highlighting promising directions for future research in anticipatory visual understanding and grounding.

\myparagraph{Future Directions.} 
Though \method achieves promising results on \task, several directions remain open for future exploration: (I) Scaling beyond the current frame limit of MLLMs to leverage richer global context. (II) Modeling transient visual cues or local information that are critical for predictive reasoning via optimizing sampling strategy. (III) Performing CoT-SFT and RL training on more RVOS datasets to improve generalization. (IV) Mitigating hallucinations during the reasoning process to improve its reliability and alignment. (V) Modeling uncertainty in future predictions to handle multiple plausible outcomes. (VI) Enabling real-time inference to support broader downstream applications.

\myparagraph{Acknowledgements} This work was supported by the National Natural Science Foundation of China (NSFC) under Grant No. 62472104 and the Science and Technology Commission of Shanghai Municipality under Grant No.~25511103600.

%
%
\bibliographystyle{splncs04}
\bibliography{main}
\end{document}